# Evaluating computational models of explanation using human judgments

Michael Pacer   Joseph Williams   Xi Chen   Tania Lombrozo   Thomas L. Griffiths
{`mpacer, joseph_williams, c.xi, lombrozo, tom_griffiths`}@berkeley.edu
Department of Psychology, University of California at Berkeley, Berkeley, CA, USA

## Abstract

We evaluate four computational models of explanation in Bayesian networks by comparing model predictions to human judgments. In two experiments, we present human participants with causal structures for which the models make divergent predictions and either solicit the best explanation for an observed event (Experiment 1) or have participants rate provided explanations for an observed event (Experiment 2). Across two versions of two causal structures and across both experiments, we find that the Causal Explanation Tree and Most Relevant Explanation models provide better fits to human data than either Most Probable Explanation or Explanation Tree models. We identify strengths and shortcomings of these models and what they can reveal about human explanation. We conclude by suggesting the value of pursuing computational and psychological investigations of explanation in parallel.

## 1 Introduction

Representing statistical dependencies and causal relationships is important for supporting intelligent decision-making and action – be it executed by human or machine. Causal knowledge not only allows predictions about what will happen, but is also used in explanations for events that have already occurred. For example, a set of symptoms might be explained by appeal to a particular disease, or an electrical circuit failure by appeal to a set of faulty gates.

Previous work in machine learning has provided a range of models for what counts as an explanation in cases involving a known causal system and observed effects.These models differ in what they allow as potential explanations or 'hypotheses' as well as in the objective function they aim to maximize (for a review, see Lacave and Díez [2002]). For example, one approach says hypotheses are settings for all unknown variables where you then choose the hypothesis that maximizes a posteriori probability given observed data [Pearl, 1988]; another allows hypotheses to be any non-empty variable setting and selects the hypothesis that maximizes the probability of observations under that hypothesis relative to every other hypothesis [Yuan and Lu, 2007]. While these models differ in their formal properties, arguments for one model over another typically come down to which provides a better fit to researchers' intuitions about the best explanations in a given case.

In this paper we evaluate four formal models of explanation by empirically investigating their fit to human judgments. Our aims are threefold. First, methods from cognitive psychology allow us to test how well competing models correspond to general human intuitions, rather than the intuitions of a small group of researchers. Second, by using human judgment as a constraint on formal models of explanation, we increase the odds of choosing an objective function with interesting properties for learning and inference. A growing literature in psychology and cognitive science suggests that generating and evaluating explanations plays a key role in learning and inference for both children and adults (for a review, see Lombrozo [2012]), so effectively mimicking these effects of explanation in formal systems is a promising step towards closing the gap between human and machine performance on challenging inductive problems. Finally, formal models of explanation that successfully correspond to human judgment can contribute to the psychological study of explanation, as almost no formal models of explanation generation or evaluation have been proposed within the psychological sciences.

We present two experiments in which we gave people information about a causal system and had them either generate explanations (Experiment 1) or eval-uate explanations (Experiment 2). The causal systems can be formally defined by Bayesian networks and correspond to those used in prior work to differentiate among models of explanation [Nielsen et al., 2008, Yuan and Lu, 2007]. Across two versions of two causal structures and across both experiments, we find that the Causal Explanation Tree [Nielsen et al., 2008] and Most Relevant Explanation [Yuan and Lu, 2007] models provide better fits to human data than either Most Probable Explanation [Pearl, 1988] or Explanation Tree models [Flores et al., 2005]. The results of our experiments identify strengths and shortcomings of these models, ultimately suggesting that human explanation is poorly characterized by models that emphasize only maximizing posterior probability.

## 2 Bayesian networks

A Bayesian network provides a compact representation for the joint probability of a set of random variables, $\mathcal{X}$, which explicitly represents various conditional independence statements between variables in $\mathcal{X}$. We specify a directed acyclic graph with a node corresponding to each variable in $\mathcal{X}$. We say that each node $X \in \mathcal{X}$ has a set of "parent nodes" (Pa($X$)), and that this gives us conditional probability distributions for every X given its parents $p(X|\text{Pa}(X))$. We assume that the full joint probability distribution can be specified this way, i.e., that $p(\mathcal{X}) = \prod_{X \in \mathcal{X}} p(X|\text{Pa}(X))$. This is equivalent to assuming that $X$ is independent of all nondescendent variables given its parents, and allows us to use the structure of the graph to read off which conditional independence relations must hold between the variables [Pearl, 1988].

Figure 1 shows an example of a Bayesian network specifying conditional probability distributions between random variables. The graph on the left (*Pearl*, named after Pearl [1988]) represents whether a particular alien has a disease ($D$), whether that alien has a genetic risk factor for that disease ($G$), and whether or not the alien was vaccinated for the disease ($V$). The graph on the right (*Circuit*) can be interpreted as a circuit that always receives input and for which we can measure the output. $A, B, C$, and $D$ are gates that, if functional, break the circuit, stopping the input from reaching the output. Each gate has an independent probability of failing and allowing current to cross through it. If the current can travel from the input to the output via any path made possible by a set of failed gates, then there will be output. These two examples hint at the richness of the Bayesian network formalism. We will continue to refer to these graphs throughout, which are the basis for our stimuli in Experiments 1 and 2, with the parameter values indicated in Figure 1.

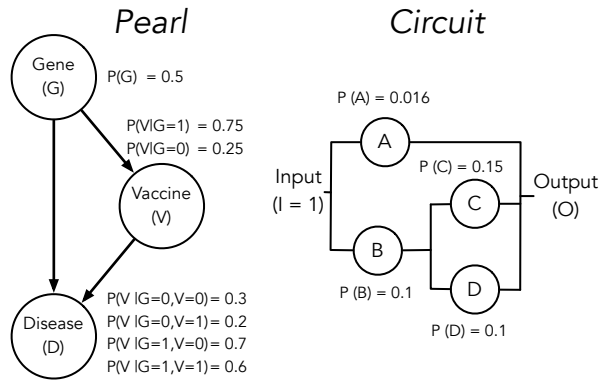

Figure 1: The *Pearl* and *Circuit* networks used in our experiments; as in Pearl [1988], Nielsen et al. [2008], Yuan and Lu [2007] and Yuan et al. [2011].

### 2.1 Explanations in Bayesian networks

Suppose we observe the values for $k$ of the variables in a graph, $\{O_1 = o_1, \ldots, O_k = o_k\}, \forall i\ O_i \in \mathcal{X}$. We may not wish to explain every observation, so let us call the variables we want to explain $O_{\text{exp}}$, with values $o_{\text{exp}}$. These values $o_{\text{exp}}$ are the "target" of our explanation, or the *explanandum*, which is a subset of $\mathcal{O}$, the set of possible observation sets. We will refer to $\hat{O}$ as the set of variables that were observed and $\hat{o}$ as the observed values. Then $O_{\text{not-exp}} = o_{\text{not-exp}}$ are those variables that are observed and unexplained (or $O_{\text{not-exp}} \equiv \hat{O} \setminus O_{\text{exp}}$).

A candidate explanation (the *explanans*, or "hypothesis") is a set of variable assignments for some of the variables not in $O_{\text{exp}}$. We exclude $O_{\text{exp}}$ to avoid circularity, though elements in $\hat{O} = \hat{o}$ but not in $O_{\text{exp}}$(i.e., observed but unexplained variables) could be included. However, we should note that most models require that every observed variable be explained; formally $\hat{O} \equiv O_{\text{exp}}$. For the sake of clarity, a hypothesis (our term for potential explanans henceforth) will be represented by $h$, the variables assigned in that hypothesis by $H$, and the set of hypotheses (treating each set of assignments as a separate hypothesis) as $\mathcal{H}$.

The first question a formal account of explanation must answer is which variables should be used in constructing $\mathcal{H}$. One possibility is for every explanation to include an assignment for every unobserved variable. However, Bayesian networks often use variables not meant to correspond to real entities in the world (e.g., a noisy-or gate for combining the influence of two causes). Additionally, there are often many variables that are not invoked in an explanation, and so a notion of "relevance" can be useful, allowing assignments to a subset of the unobserved variables (or even variables that are observed but not in $O_{exp}$).

Some models first generate $\mathcal{H}$ and then evaluate each hypothesis and rank them accordingly. Others "grow" their hypotheses by iteratively adding variables based on their ability to improve the explanation, stopping when the hypothesis cannot be improved further [Flores et al., 2005, Nielsen et al., 2008]. The hypotheses under consideration can then be evaluated and ranked, but note that what counts as an improved hypothesis and what counts as a better explanation can be based on different criteria even within the same model. Some models aim to maximize the probability of the hypothesis given the observations ($p(h|\hat{o})$) [Pearl, 1988, Shimony, 1991]. Some models are more concerned with other metrics, such as the relative likelihood of the observations under one hypothesis ($p(\hat{o}|h)$) compared to the rest of the hypothesis set [Yuan and Lu, 2007, Yuan et al., 2011]. And some models aim to maximize how much information is gained about the explanandum were the hypothesis assumed or made to be true [Flores et al., 2005, Nielsen et al., 2008].

We now introduce the four models that we consider in this paper — Most Probable Explanation [Pearl, 1988], Most Relevant Explanation [Yuan and Lu, 2007], Explanation Trees [Flores et al., 2005], and Causal Explanation Trees [Nielsen et al., 2008].

## 2.2 Most Probable Explanation (MPE)

Most Probable Explanation (**MPE**) ranks highly hypotheses with the most probable assignments to all unobserved variables, conditioning on $\hat{O}$. That is, every $h$ in $\mathcal{H}$ includes an assignment for every variable in $\mathcal{X} \setminus \hat{O}$.[1] This model leverages the intuition that the best explanation is one that is most probable given what we have observed [Pearl, 1988]. The result is

$$MPE = \arg\max_{h \in \mathcal{H}} p(h|\hat{o}). \quad (1)$$

## 2.3 Most Relevant Explanation (MRE)

Rather than choosing the hypothesis that maximizes the probability of the unobserved variables given the observed values, we could choose values for the unobserved variables to maximize the probability of the observations ($\arg\max_{h \in \mathcal{H}} p(\hat{o}|h)$). Methods that pursue this route are known as likelihood models.

One problem faced by likelihood models is that multiple hypotheses will sometimes give the same high probabilities to the observed data [Nielsen et al., 2008]. For example, consider the case where we know the structure of a causal system like the circuit in Figure 1 from Yuan and Lu [2007]. Likelihood methods would treat any hypothesis containing a union of $A$, "$B$ and $C$", or "$B$ and $D$" as equally good — the current flows equally well (perfectly), regardless of the particular path it takes. This can make it difficult to choose between these explanations within the likelihood framework.

---

[1] We might allow $h \in \mathcal{H}$ to include only those variables that are relevant for explaining $\hat{O}$. This is known instead as the maximum a posteriori model. There are a variety of possible relevance criteria as explored by De Campos et al. [2001], but this problem is substantially more computationally complex than **MPE**. Here, we focus on **MPE**.

Rather than maximizing the likelihood per se, we can instead choose the hypothesis, $h$, that has the highest likelihood *relative* to the summed likelihood of all the other hypotheses in $\mathcal{H}$ except for $h$:

$$\frac{p(\mathcal{O}|h)}{\sum_{h_j \neq h, h_j \in \mathcal{H}} p(\mathcal{O}|h_j)}. \quad (2)$$

Yuan and colleagues' Most Relevant Explanation (**MRE**) model [Yuan and Lu, 2007, Yuan et al., 2011] proposes that the best explanation maximizes this quantity. This term plays an important role in statistics, known as the Generalized Bayes Factor [Fitelson, 2007], and in psychology, as a measure of how *representative* some data is of a hypothesis [Tenenbaum and Griffiths, 2001, Abbott et al., 2012].

## 2.4 Tree-based models: ET and CET

The methods we have explored so far presume that you have $\mathcal{H}$ and then evaluate each hypothesis to determine which is best. However, in cases where the variable set is large, this can be difficult and computationally prohibitive. A class of *tree*-based models addresses this problem by using an iterative process for arriving at explanations. These models construct an explanation piece-wise, adding variables to the hypothesis one at a time, by choosing the best variable, assigning the variable a value and repeating until no further gains can be made. The resulting hypotheses are then evaluated based on some criteria, producing a list of explanations ranked by their goodness. Models differ in how they choose the best variable to add, how they decide to stop, and how they then evaluate the resulting hypotheses.

The Explanation Tree (**ET**) model — as proposed by Flores et al. [2005] — determines which variable carries the most information about the rest of the unknown nodes, conditioned on what is already known. In **ET** what is already known includes $\hat{O}$ and any variables included in hypotheses farther up the tree. This means that at the beginning (when the hypothesis is $\emptyset$) the model selects the node that provides the most information about the rest of the unobserved variables conditioned on $\hat{O}$. Formally, we grow $h'$ (the hypothesis up to that point) by choosing the $X_i$ as the maximum of $\sum_Y \text{INF}(X_i; Y|\hat{O}, h')$, where $Y$ is shorthand for $\mathcal{X} \setminus \{\hat{O} \cup h' \cup \{X_i\}\}$, or all of the variables not observed, included in the current hypothesis or currently under consideration, and $\text{INF}(\cdot)$ is a metric of informativeness. In our calculations we will use *mutual information* as our $\text{INF}(\cdot)$, as in Nielsen et al. [2008].[2]

Once a variable is chosen, each assignment creates a new branch, and that assignment is added to the interim hypothesis $h'$, and is effectively treated as an observed variable. The process is then repeated until adding any more variables is deemed to provide a hypothesis with a probability that is too low, as defined by parameter $\beta_{ET}$, or to carry too little information, as defined by parameter $\alpha_{\mathbf{ET}}$. This process provides multiple, mutually exclusive explanations that can vary in their complexity based on how much information the complexity buys.[3] Once these hypotheses are assembled, the model ranks the explanations by the posterior probability of each branch of the tree – i.e., how likely each hypothesis is, given the observed data.

Up to this point every model we have considered assumes the set of observed data is the data we are explaining, or $\hat{O} \equiv O_{\text{exp}}$. The **ET** model further assumes that we aim to reduce uncertainty of the entire variable set $\mathcal{X}$ in deciding which variables are ostensibly relevant to our explanandum, $O_{\text{exp}}$. However, these assumptions can be problematic. For example, in **ET**, a variable that is unrelated to $O_{\text{exp}}$ but carries a lot of information about other unknown variables may be added to the hypothesis despite its irrelevance to our explanans.

The Causal Explanatory Tree (**CET**) model introduced by Nielsen et al. [2008] addresses these weaknesses. Rather than using traditional measures of information such as mutual information, **CET** uses *causal information flow* [Ay and Polani, 2008] to decide how the tree will grow. Causal information flow uses the post-intervention distribution on nodes (as proposed in Pearl [2000]) rather than considering the joint probability distribution "as is". To extend Ay and Polani [2008]'s analogy, imagine pouring red dye into a flowing river. You could identify which way is downstream by tracking the red streak that results; if you were to pour in the dye just after a fork in the river, you would not find red dye in the other half of the fork. Now consider the case of a static, dammed river — a river that does not flow. If you poured the dye just after the fork, redness would gradually diffuse through the water, eventually reaching the other path from the fork and tinting the whole river. In this case, there is no concept of something being 'downstream'. Causal information attempts to capture the notion of 'downstream' influence that is absent in traditional mutual information.

We denote post-intervention distributions with a " ¯ " on a conditioned variable. If we have variables $W, X, Y, Z$, where we have observed $W = w$, intervened on $Z$ (giving us post-intervention values $\bar{Z} = \bar{z}$), then the causal information passed from $X$ to $Y$ is,

$$\sum_{x \in X} p(X = x | W = w, \bar{Z} = \bar{z}) \times$$
$$\sum_{y \in Y} p(Y = y | \bar{X} = \bar{x}, w, \bar{z}) \log \frac{p(y|\bar{x}, w, \bar{z})}{p(y|w, \bar{z})} . \quad (3)$$

This allows us to specifically ask the degree to which a variable ($X \equiv X_i$) influences the explained data ($Y \equiv O_{\text{exp}}$), treating the non-explained data as observed ($W \equiv O_{\text{not-exp}}$) and previous parts of the explanation as intervened on ($Z \equiv h'$). This solves the problem of distinguishing between explained and unexplained observations ($W \neq Y$). It also allows us to maximize information about the $O_{\text{exp}}$ rather than $\mathcal{X} \setminus \hat{O}$ as in **ET**. However, like **ET**, the **CET** model proposes variables iteratively, until no remaining variables add more causal information than the criterion $\alpha_{\mathbf{CET}}$. Then each branch is assigned the score $\log \left( \frac{p(O_{\text{exp}}|\bar{h}', O_{\text{not-exp}})}{p(O_{\text{exp}}|O_{\text{not-exp}})} \right)$ where $\bar{h}'$ is the total set of assigned values in a hypothesis at a branching point.

## 3 Comparing model and human judgments about explanations

We now compare the prediction of these four models against human judgments when both generating and evaluating explanations. We focus on explanations in the two Bayesian networks shown in Figure 1. The *Pearl* structure is derived and parameterized as in Nielsen et al. [2008]; the *Circuit* graph and its parameters are taken from Yuan and Lu [2007]. These networks have been used previously to distinguish between the performance of different models. Each network consists of several binary variables, prior probabilities on those variables, and relationships between variables. We consider the case where only one variable is observed, in *Pearl* $D = 1$ and in *Circuit* $O = 1$, and these act as both $\hat{O}$ and $O_{\text{exp}}$, i.e., each is the only variable we observe and explain in that structure.

The models diverge in how they rank explanations in *Pearl* and *Circuit*. In past research, the *Pearl* structure was used by Nielsen et al. [2008] to argue in favor of the **CET**, and the *Circuit* structure was used by Yuan and Lu [2007] to argue in favor of the **MRE**.[4]

---

[2]Flores et al. [2005] consider several versions of INF.

[3]Mutual exclusivity refers to the fact that once a variable is assigned, it holds through the rest of the tree.

[4]The **CET** had not been published by the writing of Yuan and Lu [2007]. Yuan et al. [2011] addresses **CET**

By drawing from distinct research lines we aim to be as fair as possible in testing the models.

In addition to being useful for distinguishing between models, these structures have properties that are particularly interesting from a psychological perspective. The *Pearl* structure includes complex causal dependencies that cannot be easily captured by the paradigms used in cognitive psychology. The *Circuit* structure contains explanations with equal (perfect) likelihoods for the observation, but which vary in the number of variables cited in the explanation. Research on people's preferences for simplicity bear on this case, which shows that people may choose an explanation with fewer causes even if it is less likely than other more complex alternatives [Lombrozo, 2007].

In the past, researchers used the match between their own explanatory intuitions and the models' predictions to provide support for their model. However, this method can be problematic: Nielsen et al. [2008] and Yuan and Lu [2007] conflict in their intuitions, leaving us in a quandary. We generalize the intuition-matching approach using two experiments in which we ask people to generate (Experiment 1) and evaluate (Experiment 2) explanations in cases formally equivalent to *Circuit* and *Pearl*. We used **MPE**, **MRE**, **ET**, and **CET** to rank the quality of various explanations, and analyze these rankings as they compare to the rankings derived from human explanations. By appealing to a wider array of human judgments we hope to extricate ourselves from this quandary.

## 4  Experiment 1: Generation

### 4.1  Participants

We recruited 188 participants through Amazon Mechanical Turk; 9.6% of those failed to complete the study, did not consent to taking the study, or did not follow the instructions, and 35.9% failed at least one explicit reading/attention check. This left 109 participants for analysis ($M(\text{age}) = 27.7$, %-Female $= 29.3\%$).

### 4.2  Materials & procedure

Participants were randomly assigned to either the *Pearl* or *Circuit* structure. They then were assigned to one of two semantically-enriched stories embodying a causal structure, involving either novel alien diseases or the ecology of lakes. For example, one of the two scenarios adapted from the *Circuit* structure taught participants about the effects of novel diseases on pro-

---

but that work involves more complicated scenarios than those considered here.

ducing a kind of fever.

For this scenario, participants received facts about the base rates of four novel diseases (corresponding to $p(A)$, $p(B)$, $p(C)$, and $p(D)$), and information allowing them to understand which diseases would produce the fever, which would only occur in the presence of two proteins X and Y. One disease (corresponding to $A$) produced both the necessary proteins and thereby caused the fever. The second disease (corresponding to $B$) produced one of these proteins, and when paired with either the third and/or the fourth diseases (i.e., $C$ or $D$) which produced the other protein, would be sufficient to cause the fever. X and Y were added to provide an intuitive mechanism outside of the domain of circuits that describes the complexities of *Circuit*'s causal relations. Probabilities were presented as frequencies (out of 1000) and act as realizations of the probabilities in the graphs in Figure 1.

In order to ensure that participants were paying attention, we asked questions that required simply reading the information off a figure (e.g., "Out of 1000, how many aliens have [disease $A$]?"). Participants who failed any comprehension questions were excluded from subsequent analyses. To ensure that participants' judgments were not limited by memory, the base rates and causal structure were available when answering these reading checks as well as during the generation portion of the experiment. Participants were asked to use the information that had been provided to write down "the **SINGLE BEST EXPLANATION**" for the observed effect (e.g., for a particular alien's fever), where "a 'single' explanation can include more than one causal factor." Participants were explicitly asked not to list multiple possible explanations, but rather to "identify the one explanation that you think is the best." This was meant to exclude what we call "disjunctive" explanations like "It was A or B and C and not D", or, formally, as $A = 1 \cup \{B = 1 \cap C = 1 \cap D = 0\}$.

### 4.3  Results and discussion

Participants' explanations were coded by an assistant blind to the authors' hypotheses. The coder's goal was to identify which variables were mentioned and what values were assigned to those variables. We excluded participants who gave a response that conflicted with our instructions, such as providing a disjunctive explanation.

In *Circuit*, most participants provided explanations that fell into one of two options: $BC$ (43%) or $A$ (40%), and, in *Pearl*, most participants chose one option: they attributed the disease to the presence of a genetic risk factor and not receiving the vaccine (73%, see Figure 1).

For the explanations participants generated, we computed measures of explanation quality under each of the four models and saw which models gave better scores to those explanations that were generated more frequently. This process provides us a rank for each participant's explanation according to each of the models and a rank of how frequently each explanation was generated, which allows us to calculate a Spearman rank-order correlation between participant's aggregate explanation choices and the models' predictions, see Table 1.

Note, we used two versions of the *tree*-algorithms: one where explanations not reached by the tree received the lowest possible rank (which we give the subscript "tree"), and one where we ignored these exclusions and applied the evaluation criteria used at each branch point. The tree models were designed to both generate and evaluate explanations "on the fly", but it is not clear whether the way models *generate* explanations has led to their success in previous literature. Model success (or failure) may be the result of the branch evaluation criterion, rather than the result of the algorithm for generating hypotheses. This is why we analyze these parts of the algorithms separately.

We find that **MRE** and **CET** are most consistent with participants' judgments (though they still only reach marginal significance in the *Circuit* case). In contrast, for both structures, models that rely only on an assignment's probability (i.e., **MPE** and **ET**) poorly predict the explanations that people generate (in *Circuit*, **MPE** had a negative coefficient).

The major weakness of the tree versions of **CET** and **ET** lies in the fact that once a node is chosen for expansion, it remains expanded. Thus, mutually exclusive explanations cannot be reached in the same tree. That is, in *Circuit*, $A$ and $BC$ were the two most popular explanations and $A \cap BC = \emptyset$, so the first step to include either $A$ or $B$ will preclude the other explanation. Empirically, participants are roughly split between these two explanations, which suggests that any method that generates a unique best explanation will always fail to capture the variability that results when people are generating explanations, even if those people are generating explanations about the same system. We studied only deterministic algorithms which may be causing the models to diverge from people in how they generate hypotheses. Adding probabilistic rules may also be important for accounting for uncertainty about the parameter estimates, which in the real world are typically not given to you but must be inferred from data as well.

Note that **CET** in this case treats all explanations that sufficiently determine the observations as having equivalent rank. Because the system is deterministic, all 38 of the sufficient explanations are ranked as number 1 — or rather, because they are so numerous, number 19. This is a problem unique to **CET**, and results from its use of intervention, which ignores variables' prior distributions in determining an explanation's score.

## 5 Experiment 2: Evaluation

In Experiment 1, we found evidence that at least some of the proposed models capture people's explanatory intuitions. Of course we should have expected some of the models to perform well; what is remarkable is how poorly some of the models did. In particular, we saw surprisingly poor performance from the tree-growth models as compared to their exhaustive-search evaluative counterparts.

Generating explanation is harder than only evaluating them — generation requires searching through the hypothesis set and then evaluating the generated explanations, while evaluation only requires computing a known evaluation function. The tree versions of the tree models are designed to make generation tractable. However, if complexity were the primary hurdle, in *Circuit* where the hypothesis space was much larger, we would expect tree methods to perform comparatively better than in *Pearl*. But they were relatively *worse*. This was due to the fact that the tree models were guaranteed to cut off at least 40% of participants since $A$ and $BC$ were the top choices, and cannot be reached in the same tree.

It is striking that methods that relied on probability (**MPE** and **ET**) performed so poorly in contrast to **MRE** and **CET**. However, these results may only apply to situations in which explanations are generated; explanations with large absolute probabilities may be difficult to access when generating explanations but could still be preferred if people only need to evaluate predefined hypotheses. There are many cases in which a hypothesis proves incredibly hard to generate, but once generated quickly becomes welcomed as the best explanation for many phenomena (e.g., Newton's and Einstein's physics). And, if conquering search problem is one of the driving factors behind the success of **MRE** and **CET**, it is possible that they could fail in the evaluation case.

Table 1: Rank-correlations for models and human data in Experiment 1, $P_{\text{val}} < 0.05$ in **bold**, $< 0.10$ in *italics*.

|  | *Circuit* |  | *Pearl* |  |
| --- | --- | --- | --- | --- |
| Models | $\rho_{\text{Spearman}}$ | $P_{\text{val}}$ | $\rho_{\text{Spearman}}$ | $P_{\text{val}}$ |
| **MPE** | -0.06 | 0.631 | 0.32 | 0.449 |
| **MRE** | *0.20* | *0.074* | **0.83** | **0.017** |
| **ET** | 0.08 | 0.460 | 0.17 | 0.700 |
| **ET**$_{\text{tree}}$ | 0.01 | 0.900 | 0.41 | 0.310 |
| **CET** | *0.22* | *0.055* | **0.93** | **0.003** |
| **CET**$_{\text{tree}}$ | 0.06 | 0.590 | **0.77** | **0.032** |

In order to test these ideas, we conduct an experiment that is almost identical to Experiment 1. But, rather than asking people to generate explanations, we take that burden off of their shoulders. Instead, we ask them to evaluate a set of explanations that we generate for them.

### 5.1 Participants

A total of 245 participants were recruited through Amazon Mechanical Turk, with 9.8% excluded for failing to provide consent or otherwise complete the study and 25.3% excluded for failing one or more reading checks. This left 165 participants for analysis ($M(\text{age}) = 31.3, \%$-Female $= 34\%$): 46 in the disease version of *Circuit*, 46 in the lake version of *Circuit*, 34 in the disease version of *Pearl*, and 39 in the lake version of *Pearl*.

### 5.2 Stimuli

An explanation was included in the study if either criterion held:

- The explanation was generated by more than one participant in any one condition in Experiment 1.
- The explanation was in the top two explanations generated by any of the models.[5]

This yielded thirteen explanations for the *Circuit* causal structure and six for the *Pearl* causal structure.

### 5.3 Procedure

The materials and methods were nearly identical to those in Experiment 1, with the following important change: instead of providing an explanation, participants were asked to rate the quality of several provided explanations. Specifically, they were asked to rate each explanation "by placing the slider next to each explanation along the spectrum from Very Bad Explanation (furthest to the left) to Very Good Explanation (furthest to the right)," where intermediate ratings could fall anywhere in between.

---

[5]Because there are many ways one can interpret what counts as one of the two "top" explanations, we allowed the top two as defined by *any* interpretation found in the literature of how to rank a model's results. For example, Yuan [2009] and Yuan et al. [2011] include only minimal explanations (i.e., explanations for which no subset has appeared prior to it in the ranking of explanations) when determining the results of **MRE**, whereas Nielsen et al. [2008] simply listed explanations based on their scores regardless of their minimal or non-minimal status.

Although the sliders were not presented with a numbering, positions implicitly corresponded to values between 0 and 100. Based on these ratings we can again create an explanation ranking for each participant, with ties being treated as in Experiment 1 as a repeated average value. By using ranks rather than continuous ratings we need only assume that participants have a monotonic relationship between bad and good, and avoid making assumptions about the particular nature of that scale for each participant.

### 5.4 Assessing model predictions

For each model, we calculated the scores assigned to the explanations that were provided to human participants. Because we were interested in explanation evaluation, we did not limit the ranks derived from **CET** or **ET** to those generated by the trees, but we did limit **MPE** to complete assignments, as otherwise it would be equivalent to **ET**.

To generate scores indicating the quality of each model, we created a set of intersection proportions. To illustrate, were we to consider only a single participant, this involves the following process. We take the human ranking as the veridical ranking. We then check whether the model's top rank explanation is the same as the participant's. We then check whether the model's two highest-ranked explanations are included in either of the two highest-ranked human explanations. We continue to do this for the whole explanation set, identifying the number of model explanations that were ranked at a level less than or equal to each level of human ranking. We can repeat this with every participant, to obtain the number of explanations matched at each rank for each participant. We can then take the average of these scores at each rank, giving us the intersection size for the full population.

It is important to note that the absolute intersection size is less useful than the proportion when we are comparing between causal structures. We can transform these values into intersection proportions by dividing each value by the total number of model explanations. This maps to a measure of how many of the model's top explanations are thought by the models to be at least as good as those generated by the average person up to that point.

To illustrate, suppose that we had explanation set $\mathcal{H} : A$, $BC$, $BD$, $ABCD$, and $B$, and we were con-sidering a participant($P$) with a ranking of $P(1) = BC$, $P(2) = A$, $P(3) = ABCD$, $P(4) = BD$, and $P(5) = B$. To compute a model's performance, we would look at the ranking that the model($M$) assigned to the different explanations. If their top ranks matched, i.e., $M(1) = BC$ was the model's top choice, then the first value would be $V(M, P, 1) = \frac{1}{5} = \frac{|M(1)=\{BC\} \cap P(1)=\{BC\}|}{|\mathcal{H}|}$, and if it was not the score would be 0 since $M(1) \cap P(1) = \emptyset$. This process would be repeated for the first and second values, i.e., the next value is $V(M, P, 2) = \frac{|\{M(1)M(2)\} \cap \{P(1)P(2)\}|}{|\mathcal{H}|}$, and so on until we got to $V(M, P, 5)$ which will necessarily equal 1 since both rankings were defined relative to the same set, meaning the two sets are equivalent and are also both equivalent to $\mathcal{H}$.

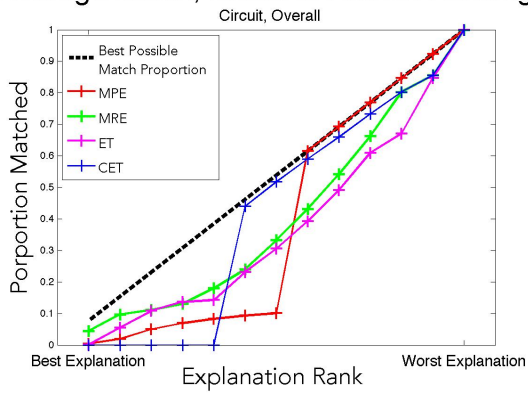

Figure 2: Results for Experiment 2: Average intersection proportions for *Circuit* conditions.

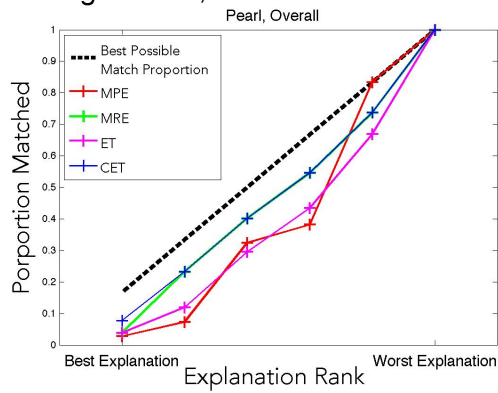

Figure 3: Results for Experiment 2: Average intersection proportions for *Pearl* conditions.

Figure 2 displays the intersection proportion for the *Circuit* structure, and Figure 3 displays those for the *Pearl* structure.

Another method for capturing overall model performance is to take the sum of the average values at each point. The best one can do in the intersection proportion is to match every explanation up to that rank at each rank. A perfect summary score is, $\sum_{i=1}^{|\mathcal{H}|} i/|\mathcal{H}|$. For *Circuit* the maximum summed intersection value is $\sum_{i=1}^{13} i/13 = 7$ and for *Pearl* it is $\sum_{i=1}^{6} i/6 = 3.5$.[6] These values can be found in Table 2.

### 5.5 Results and discussion

As you can see in Figures 2 and 3, both **MRE** and **CET** are closer to the dotted line in general, i.e., they are better on average than either **MPE** or **ET**.

One interesting pattern to note is a trend that echoes results for **CET** in Experiment 1. **CET** stays flat at zero for a while and then rapidly accelerates as it goes forward. This is a consequence of the interaction between **CET**'s reliance on intervention and the deterministic causal system in the *Circuit* condition. Because so many of the explanations are sufficient for bringing about the effect in question, many explanations share the role of the 'best' explanation. And because we choose an explanation's rank in the case of a tie as the average rank of all those in the tie had they not been in a tie, many of the best explanations are given a fairly high value. Thus, once we get to the sixth item, $M(1)$-$M(5)$ have had equal scores to $M(6)$, and once the values pass that threshold **CET**'s $V(M, P, \cdot)$ rapidly catches up to and passes **MRE**'s (which was otherwise in the lead). **MPE**, on the other hand, has the opposite problem: only two of its values are defined and so the other eleven explanations all receive a score of 8, resulting in perfect performance from 8 onwards (though most of its ranks are, by definition, undefined).

---

[6]One could think of this as an estimate of the area under the curve defined by the intersection proportions.

Table 2 shows that in both structures **CET** does the best, followed by **MRE**, then **MPE** and finally **ET**.

## 6 General discussion

We began this paper with the aim of systematically evaluating formal models of explanation against human intuitions as well as clarifying human explana-tion through the lens of computational models. We consider how our results address these aims.

Table 2: Summed intersection values for models.

| Models | *Circuit* Score | *Pearl* Score |
|---|---|---|
| **MPE** | 5.26 | 2.64 |
| **MRE** | 5.43 | 2.96 |
| **ET** | 4.99 | 2.55 |
| **CET** | **5.60** | **3.00** |
| Max Value: | 7 | 3.5 |

## 6.1 Evaluating models of explanation

We find that **CET** and **MRE** provide reasonable but imperfect fits to human judgments in both the *Circuit* and *Pearl* structures, and for both explanation generation and evaluation. **MPE** and **ET** perform less well. This suggests that human explanation is not explained well by appealing to maximum posterior probability values. Instead, it seems that a measure of evidence (**MRE**) or causal information (**CET**) may better model human explanation.

These findings indicate that the algorithms used for generating explanations in the tree methods (**ET** and **CET**) fail to capture an important aspect of human intuitions about explanation — explanations that are radically different from one another (i.e., that cannot be reached by the same tree) may both be seen as valid explanations. In the generation task, the purely evaluative tree models outperformed their generative counterparts. The evaluation function seems to be quite important, but it has been emphasized less than the generation algorithm in previous work [Flores et al., 2005, Nielsen et al., 2008]. The evaluation function merits closer inspection.

Speaking generally, our work reveals the degree to which a model's objective alters that model's predictions. Our analyses highlight the problem with using hard intervention in deterministic cases. **CET** gave the same score to all 38 sufficient explanations that, presumably, we would want the model to distinguish. **MPE** and **ET** excel at doing what they were created to do, but we may wish to distinguish between their goals (which do not correspond closely to human explanation judgments) and the goals of models like **CET** and **MRE** (which do).

## 6.2 Bidirectional implications from human and formal explanation

These results indicate that formally characterizing the objective function implicit in human explanation may be a challenging but exceptionally useful task. The variability in how well these formal models performed demonstrates that despite seeming straightforward, how people choose a good explanation has many hidden subtleties and complexities. The good performance of **CET** and **MRE** relative to **MPE** and **ET** suggest that human explanation is likely more concerned with causal intervention or the relative quality of a hypothesis than it is with absolute judgments of posterior probability. But the alternative hypothesis set and the role of intervention have received relatively little attention in psychological research on explanation. On the other hand, simplicity was not explicitly represented in the formal models we explored (but, see De Campos et al. [2001]), but has been found to affect human explanatory judgments [Lombrozo, 2012]. Then, it is surprising that a large proportion of people explain using $BC$ over $A$ in the *Circuit* example, when $BC$ is both less likely and more complex than $A$. Probability, simplicity, intervention and alternative hypotheses seem to weave a rather complex image — an image just asking to be unraveled.

All of the models we studied require knowing a priori the causal structure and parameterization, whereas people must infer these values from finite amounts of data. Though explanation has been tied to improved learning, we know much less about how the learning process and the processes for generating and evaluating explanations interact with one another. Additionally, developing extensions of these models that can learn from finite amounts of data will increase the expressivity of the models while also making them more able to deal with the problems that both humans and many real intelligent systems face.

## 6.3 Conclusion

Given that explanation plays an important role in human inductive judgments [Lombrozo, 2012], where humans still outperform artificial systems, we propose that models will benefit from a closer match to human judgments. And conversely, given that formal models need to make explicit the roles played by different parts of the explanatory problem and its solution, we propose that psychological accounts of explanation will benefit from models that precisely specify formal characteristics for what makes a good explanation. Both inquiries benefit from attending to the other. Our work, in simultaneously analyzing models of explanation from artificial intelligence and the psychology of human explanation, embodies this view.

### Acknowledgements

This work was supported by a National Defense Science and Engineering Graduate Fellowship and a Berkeley Fellowship awarded to MP, grant number DRL-1056712 from the National Science Foundation to TL, and grant number IIS-0845410 from the National Science Foundation to TLG.